%% file: 00-main.tex
\pdfoutput=1

\documentclass[11pt,a4paper]{article}
\usepackage[hyperref]{emnlp2021}
\usepackage{times}
\usepackage{latexsym}

\usepackage{times}
\usepackage{latexsym}

\usepackage[T1]{fontenc}

\usepackage[utf8]{inputenc}

\usepackage{microtype}

\usepackage{graphicx}
\usepackage{tabularx}
\usepackage{amssymb}
\usepackage{multirow}
\graphicspath{{images/}}

\title{Perturbing Inputs for Fragile Interpretations in Deep Natural Language Processing}

\author{Sanchit Sinha, Hanjie Chen, Arshdeep Sekhon, Yangfeng Ji, Yanjun Qi \\
Department of Computer Science \\ University of Virginia \\ Charlottesville, VA, USA \\ \texttt{\{ss7mu,hc9mx,as5cu,yangfeng,yq2h\}@virginia.edu}}

\input{preambular}
\begin{document}

\maketitle
\input{001-abstract}

\input{01-introduction}

\input{02-related_works}
\input{03-method}

\input{03-diagram-figure}
\input{031-algorithm}

\section{Experiments}
\input{051-underHood}

\input{05-evaluation}
\input{060-table-main}
\input{04-experiments}

\input{06-results}

\input{064-quality}
\input{065-quality-tables}

\input{07-conclusions}

\clearpage
\bibliographystyle{acl_natbib}
\bibliography{refAE,refNLP}

\clearpage
\appendix
\section{Appendix}
\label{sec:appendix}
\input{062-baseline2appx}
\input{063-tables2Appx}
\input{063-violin2appx}
\input{99-visual}


\end{document}

%% file: preambular.tex
\usepackage{amsmath}
\usepackage{amsfonts}
\usepackage{amssymb}
\usepackage{color}
\usepackage{url}
\usepackage[boxed,noend]{algorithm2e}

\usepackage{flushend}

\newcommand{\sref}[1]{Sec.~\ref{#1}}

 \newcommand{\bit}{\begin{itemize}}
 \newcommand{\eit}{\end{itemize}}
 \newcommand{\ben}{\begin{enumerate}}
 \newcommand{\een}{\end{enumerate}}

\newcommand{\method}{``ExplainFooler''\xspace}

\newcommand{\victimI}{\textsc{Integrated Gradient}\xspace}
\newcommand{\victimL}{\textsc{LIME}\xspace}
\newcommand{\dnlp}{deep NLP\xspace}
\newcommand{\nobj}{``Location of Mass (LOM)''\xspace}
\newcommand{\diff}{``L2 Norm''\xspace}
\newcommand{\nobjs}{LOM\xspace}
\newcommand{\dnobjs}{``Delta LOM``\xspace}

\newcommand{\diffs}{L2 Norm\xspace}

\newcommand{\qnote}[1]{\textcolor{blue}{QNOTE:#1}}
 \newcommand{\hnote}[1]{}

\def\x{{\mathbf x}}

\usepackage{enumitem}

\usepackage{titlesec}

\titlespacing\subsubsection{0pt}{5pt plus 0pt minus 1pt}{1pt plus 0pt minus 0pt}
\titlespacing{\paragraph}{0pt}{3pt}{3pt}[3pt]

\usepackage{etoolbox}
\makeatletter
\preto{\@tabular}{\parskip=3pt}
\makeatother

\setlist[itemize]{leftmargin=*}

%% file: 001-abstract.tex
\begin{abstract}
Interpretability methods like \victimI and \victimL are popular choices for explaining natural language model predictions with relative word importance scores. These interpretations need to be robust for trustworthy NLP applications in high-stake areas like medicine or finance. Our paper demonstrates how interpretations can be manipulated by making simple word perturbations on an input text. Via a small portion of word-level swaps, these adversarial perturbations aim to make the resulting text semantically and spatially similar to its seed input (therefore sharing similar interpretations). Simultaneously,  the generated examples achieve the same prediction label as the seed yet are given a substantially different explanation by the interpretation methods. Our experiments generate fragile interpretations to attack two SOTA interpretation methods, across three popular Transformer models and on two different NLP datasets. We observe that the rank order correlation drops by over 20\% when less than 10\% of words are perturbed on average. Further, rank-order correlation keeps decreasing as more words get perturbed. Furthermore, we demonstrate that candidates generated from our method have good quality metrics. Our code is available at: \url{github.com/QData/TextAttack-Fragile-Interpretations}.
\end{abstract}

%% file: 01-introduction.tex
\section{Introduction}
Recently, the use of natural language processing (NLP) has gained popularity in many security-relevant tasks like fake news identification \cite{zhou2019fake}, authorship identification \cite{okuno2014challenge}, toxic content detection~\cite{perspectiveapi}, and for text-based automated privacy policy understanding~\cite{polisis18}. Since interpretations of NLP predictions have become necessary building blocks of the SOTA \dnlp{} workflow, explanations have the potential to mislead human users into trusting a problematic interpretation. However, there has been little analysis of the reliability and robustness of the explanation techniques, especially in high-stake settings, making their utility for critical applications unclear. 

Research has shown that it is possible to disrupt and even manipulate interpretations in deep neural networks \cite{Ghorbani_Abid_Zou_2019,NEURIPS2019_bb836c01}. The core idea in this literature centers around ``fragile interpretations''. \cite{Ghorbani_Abid_Zou_2019} defined that an interpretation is \textit{fragile} if, for a given  input, it is possible to generate perturbed input that achieves the same prediction label as the seed, yet is given a substantially different interpretation. Fragility limits how much we can trust and learn from specific interpretations. An adversary for ``fragile interpretations'' could  manipulate the input to draw attention away from relevant words or onto desired features. Such input manipulation might be especially hard to detect because the actual labels have not changed. 

The literature includes two relevant groups: (1) to conduct model manipulations \cite{slack2019can,wang-etal-2020-gradient} (details in Sec.~\ref{sec:related}), and (2) to manipulate input samples \cite{Ghorbani_Abid_Zou_2019}.  There has been little attention studying fragile interpretations via input manipulation in \dnlp{}.

\input{03-motivation-figure}



In this paper, we propose a simple algorithm \method that can make small adversarial perturbations on text inputs and demonstrate fragility of interpretations. We focus on optimizing two objective metrics - \diff or a proposed \dnobjs,  searching for small word-swap-based input manipulation to produce misleading interpretations and using semantic-oriented constraints to constrain the manipulations. Figure~\ref{fig:anotherexample} provides one example perturbation process. In summary, this paper provides the following contributions:
\begin{itemize}[noitemsep,topsep=2pt]
    \item Our input perturbation optimizes to increase the objective metric (\diff or \dnobjs) that measures difference between the original and generated interpretations. The \nobjs score captures the approximate center ``position'' of an interpretation and  summarizes it to a scalar. 

    \item We propose an effective algorithm \method to optimize the objective metric via an iterative procedure. Our algorithm generates a series of increasingly perturbed text inputs such that their explanations are significantly different from the original but preserving predictions. 

    \item Empirically, we show that it is possible to find perturbed text examples to fool interpretations by \victimI and \victimL, even on NLP models that are relatively more robust. 
\end{itemize}
The approximate process and results of word perturbation using our approach is detailed in Figure~\ref{fig:motivation}.

%% file: 03-motivation-figure.tex
\begin{figure}[t]
    \begin{center}
     \includegraphics[width=0.4\textwidth]{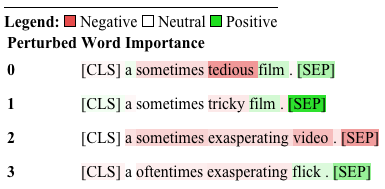} 
    \caption{The figure demonstrates the input perturbation process for an increasing number (levels) of word perturbations. The red color depicts negative attribution, and the green shows positive attribution. The saturation of the colors signifies the magnitude of the said attributions. Note: the interpretations gradually become more and more different from the original, although the semantic meaning of the sentence does not change drastically. The model still predicts the correct original output, but the interpretations become senseless as more words get perturbed.[Example taken from the SST-2 dataset. Interpretations calculated using Integrated Gradients on DistilBERT model. Best viewed in color]
\label{fig:motivation}  }
    \end{center}
    \vspace{-2mm}
\end{figure}

%% file: 02-related_works.tex
\section{Related Work}
\label{sec:related}

\paragraph{Interpretation Methods: } Several interpretation methods have been proposed\cite{shrikumar2017learning, li2015visualizing, bach2015pixel, shrikumar2017learning} to calculate feature importance scores. Two well-known methods in this area are Integrated Gradients (IG) \cite{sundararajan2017axiomatic} and Local Interpretable Model Explanations (LIME)\cite{ribeiro2016should}. IG computes the scores by summing up the gradients along a path from the baseline to the input in a fixed number of steps and subsequently multiplied by the input itself. IG overcomes the saturation problem discussed in \cite{shrikumar2017learning,sundararajan2017axiomatic}. On the other hand, LIME is a completely black-box approach which explains the predictions of any classifier in an interpretable and faithful manner, by learning an interpretable model locally around the prediction by training the model on perturbations generated around the input.

\paragraph{Fragile Interpretations}
More recently, several works have focused on discussing the robustness of the said interpretations. Studies have demonstrated that the interpretations generated are not robust and can be easily manipulated due to high dimensionality of networks. \cite{Ghorbani_Abid_Zou_2019,NEURIPS2019_bb836c01,slack2019can,wang-etal-2020-gradient}. Multiple other works have tried to fix the problem by making interpretations robust \cite{pmlr-v119-lakkaraju20a,rieger2020simple}.

\cite{wang-etal-2020-gradient} demonstrated that it is possible to introduce a new model over the original and alter gradients, to fool  gradient-based interpretation methods. Similarly, \cite{slack2019can} showed that black-box interpretation methods can also be fooled by allowing an adversarial classifier component. More recently, \cite{zafar2021lack} demonstrated empirically. that interpretability methods produce varying results on the same models but differently initialized.

\paragraph{Adversarial Examples that fool NLP Predictions:}  Adversarial examples are inputs to a predictive machine learning model that are maliciously designed to fool the model predictions \cite{goodfellow2014explaining}. Multiple recent works have focused on applying the concept of adversarial examples on language inputs, including (1) attacks by Character Substitution \cite{ebrahimi2017hotflip, gao2018black, li2018textbugger}; (2) attacks by Paraphrase \cite{SEARS-Ribeiro2018-ue, Iyyer18Paraphrase}; (c) attacks by  Synonym Substitution \cite{alzantot2018generating,Jin_Jin_Zhou_Szolovits_2020,Kuleshov2018AdversarialEF, papernot2016crafting}; (d) attacks by Word Insertion or Removal \cite{liang2017deep,samanta2017towards};  (e) attacks by limiting $L_p$ distance in a latent embedding space \cite{zhao2017generating}.  
Our proposed algorithm is closely connected to the TextFooler algorithm  \cite{Jin_Jin_Zhou_Szolovits_2020} that searches for input perturbations to achieve mis-classification. Differently, we optimize the \diff and \nobj objective directly on the input space for fragile explanations. 

%% file: 03-method.tex
\section{Proposed Method}


In this section, we present our algorithm to generate perturbed sentences that demonstrate fragile interpretations. First, we propose the metric \nobj and \diffs, followed by a discussion on the search strategy to optimize the objective metrics. Subsequently, we discuss the interpretation method choices and end with the final candidates' selection procedure and pseudo-code for our algorithm (Algorithm\ref{algo:algofig}). We denote a text input as $\x$ and its word importance score vector (from a specific interpretation strategy on a particular NLP model) using notation $I$.

\subsection{Difference Metrics on Interpretation}
\label{sec:diff-metrics}
To quantify the difference between two interpretations, we propose two objective metrics - \dnobjs and \diff. These metrics are divergent - that is higher the metric, the more different the interpretations.

\subsubsection{\nobj Score}\
First, we propose a metric inspired by \cite{Ghorbani_Abid_Zou_2019} which provides a quantifiable ``position'' of the interpretations of a sentence. First, we  define the \nobj score as:
\begin{equation}
    \nobjs(I) = \frac{\sum_{t=0}^{t=n-1} (i_{t} * t)}{\sum_{t=0}^{t=n-1}i_{t}}
\end{equation}
Here $n$ is the length of the sentence (along with starting/end special tokens). And $i_t$ is the interpretability score assigned to the token at index `t'. We  then propose to calculate the \dnobjs metric as: the difference between the \nobjs scores on the two interpretations $I_1$ and $I_2$:
\begin{equation}
    \Delta \nobjs(I_1,I_2) = | \nobjs(I_1) - \nobjs(I_2) | 
\end{equation}
The intuition behind this metric comes from the fact that changing the approximate position of the ``center'' of interpretations changes the relative position and magnitudes of interpretations. This observation is demonstrated in Figure \ref{fig:anotherexample}.

\subsubsection{L2 Norm Metric}
We also propose to use a standard \diffs  to measure difference between two interpretations. Mathematically it is computed as follows:
\begin{equation}
    \diffs (I_1,I_2) = \| I_1 - I_2\|_2 
\end{equation}
\diffs quantifies the extent of difference, higher the \diffs - higher the difference in pattern of two  interpretations. 

\subsection{Searching for Word-level Perturbations}
Our objective is to perturb a seed input $\x$, into a slightly-modified text  $\x_{adv}$, so that $\Delta \nobjs$ or $\diffs$ is maximized under a set of constraints.


First, we rank each word of an input sentence in the order of their importance to a model's predictions. This is done by the Leave-one-out approach \cite{li2016understanding}, which removes each word from the sentence one at a time and measures the change in prediction values, ranking the words which produce the greatest change as most important. Subsequently, we start our search in decreasing order of word importance and substituting each word with their $k$ closest nearest-neighbors according to their counter-fitting synonym embeddings \cite{Mrksic2016CounterfittingWV}. For every subsequent word replacement, interpretation is calculated according to victim interpretation strategy we try to attack.

\subsection{Ensuring Constraints}
\label{sec:constraints}
We enforce the following four constraints for each perturbed candidate to ensure candidates do not lose their linguistic structure and approximate semantic meaning of the seed input.
\begin{itemize}[noitemsep,topsep=3pt]
    \item Repeat Modification: Stops the same word from getting perturbed more than once.
    \item Stop Word Modification: This excludes pre-defined stop words from getting perturbed.
    \item Word Embedding Distance: Swaps the original word with words that have less than a particular embedding distance using Counter-Fitting Embeddings.
    \item Part of Speech: Replaces the original word with only words from the same part of speech.
    \item Sentence Embedding: Ensure the difference in the Universal Sentence Embedding is less than a pre-defined threshold \cite{Cer18USE}.
\end{itemize}

\subsection{Victim Interpretation Choices}
\label{sec:calcinterpret}
\paragraph{Integrated Gradient:}
We calculate \victimI \cite{sundararajan2017axiomatic} interpretations of NLP models using the open-source package Captum \cite{kokhlikyan2020captum} that provides accurate implementations of various interpretation methods. We use the popular \victimI algorithm to calculate the importance scores on the embedding space of the models. Once the interpretations are calculated, they are summed up along the dimension axis to derive the word importance scores. Subsequently, the $\Delta$\nobjs and \diffs scores of each candidate perturbation are calculated against the original input's interpretation.
\paragraph{LIME:}
The \victimL interpretations are calculated using the official LIME code provided by \cite{ribeiro2016model}. We normalize the \victimL scores by dividing the vector with its $L_2$-norm. Subsequently, the $\Delta$\nobjs and \diffs scores of each candidate perturbation are calculated against the original input's interpretation.

\subsection{Finding the ideal candidate}
Once we obtain all the candidates and their metric scores on every candidate achieving the same prediction label as the original, we store those ideal candidates with each `$m$' number of words perturbed. This gives us a list of candidates for each level of word perturbation and the associated change in objective metric scores. Next, for each level, the candidate with the highest metric score against the original is chosen. Finally, we convert the number of perturbed words into a ratio with respect to the input's length. This is done to take into account the varying sentence lengths and get a normalized measure. The ratio is limited to 50\% because once more than half the words are perturbed, the sentence starts losing its semantic meaning. The complete selection process is schematically detailed in Figure~\ref{fig:schematic}














%% file: 03-diagram-figure.tex
\begin{figure}[th]
\vspace{-1mm}
     \includegraphics[width=0.5\textwidth]{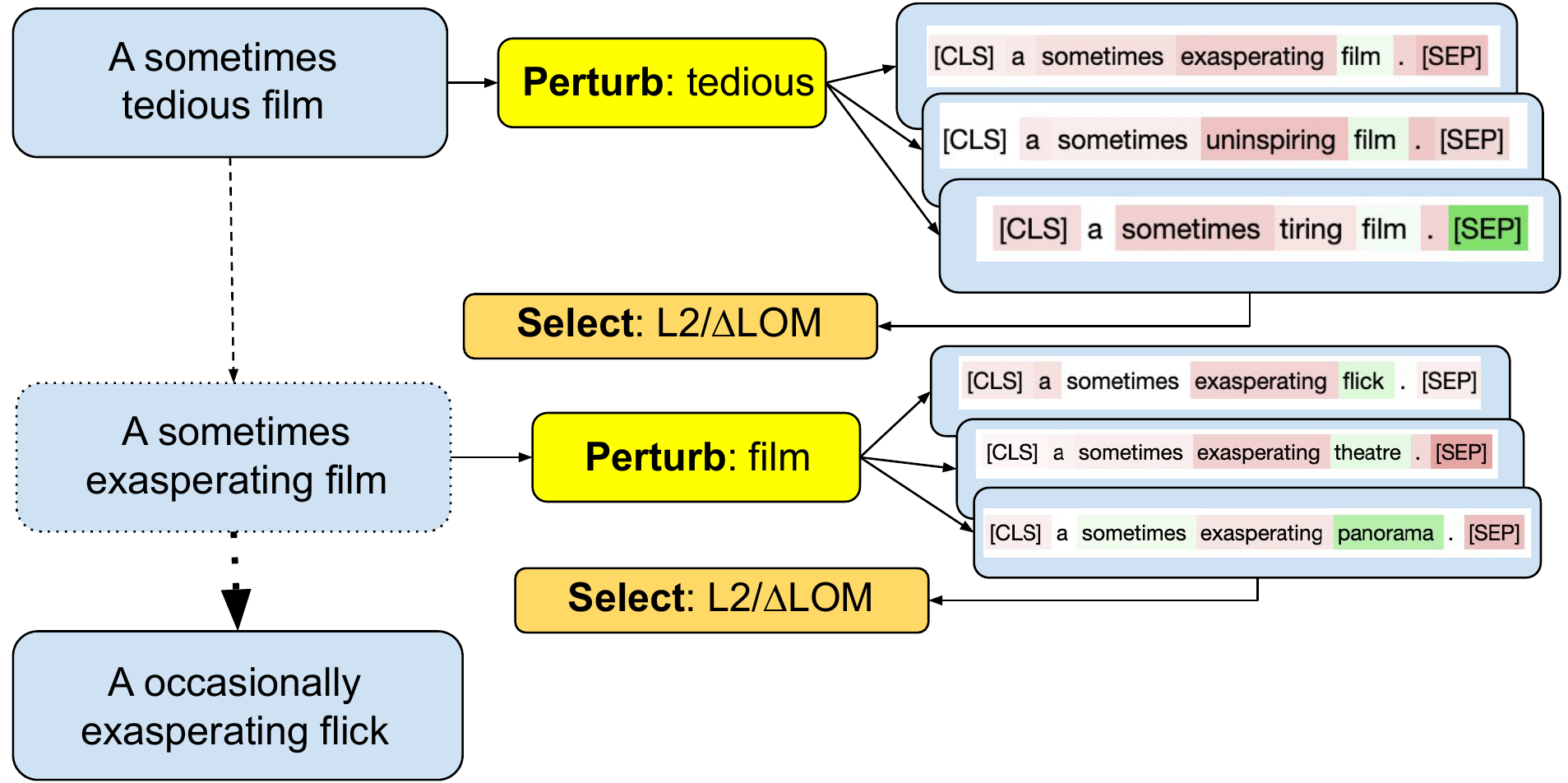}
        \caption{A schematic diagram of the proposed \method algorithm. In the figure, the ``Perturb'' step generates a list of all possible perturbations according to the constraints as discussed in \sref{sec:constraints}. The interpretation are generated as discussed in \sref{sec:calcinterpret}. The selection process uses objective metrics explained by \sref{sec:diff-metrics}.\label{fig:schematic}}

        \vspace{-5mm}
\end{figure}

%% file: 031-algorithm.tex
\subsection{Algorithm}

Algorithm \ref{algo:algofig} \method provides pseudocode to compute and select a list of  candidates that can induce fragile explanations. 
Our implementation adapts and builds on top of the open-source package TextAttack \cite{morris2020textattack}.  
\begin{algorithm}
\DontPrintSemicolon
\KwResult{A - list of candidate sentences ordered by number of words perturbed from original }
For each sentence in dataset\;
 $A\gets$empty\;
 $S\gets$original sentence\;
 $I_0\gets$ InterpretMethod(S)\;
 $P\gets$ordered list of important words (LOO)\;
\While{$<=$50\% of words perturbed from P}{
 $w\gets P[0]$\;
 $C\gets$empty\;
\While{Possible perturbations exist}{
 $c\gets$Perturb $S$ and get candidate\;
\eIf {constraints pass \textbf{and} prediction label is same as S}
{ $I\gets$ InterpretMethod(c)\;
 $\Delta diff\gets diff(I_0,I)$,
 $C\gets C \cup (\Delta diff,c)$\;}
  {continue}
}
 $A\gets A$ $\cup$ c where max(diff)\;
$P\gets$ remove $P[0] $\;
 }
\caption{The \method algorithm\label{algo:algofig}}

\end{algorithm}

\begin{figure*}[!ht]
    \begin{center}
     \includegraphics[width=0.9\textwidth]{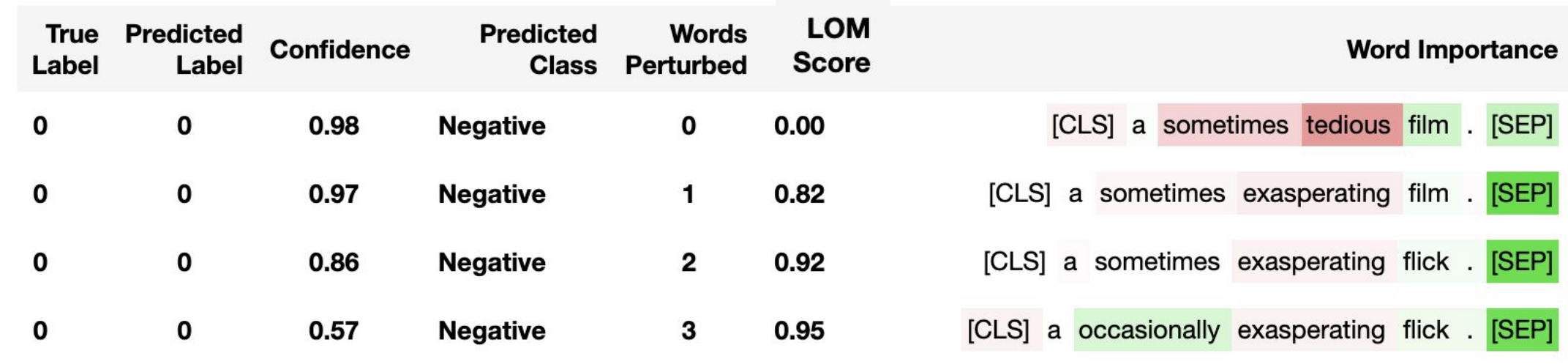} \caption{The figure demonstrates the $\Delta$\nobjs score for an increasing number of word perturbations. The interpretations gradually become more and more different from the original although the semantic meaning of the sentence does not change drastically. We can see that the model still predicts the original output but the interpretations become senseless as the $\Delta$\nobjs score increases.[Best viewed in color]
          \label{fig:anotherexample} }
    \end{center}
\end{figure*}

%% file: 051-underHood.tex
\subsection{Data Summary}
\label{sec:datasummary}
The experiments are conducted on three different datasets for text classification task. Experiments are conducted on the validation set for SST-2\cite{socher2013recursive}, test set for AG News\cite{zhang2015character} and test set for IMDB dataset\cite{maas2011learning}. We select the first 500 sentences from the SST-2 and AGNews datasets and 100 sentences from the test set from IMDB dataset to run our experiments. We discard sentences with just 2 words or less.
\begin{itemize}[noitemsep,topsep=0pt]
\item \textbf{SST-2}: The Stanford Sentiment Treebank-2 dataset for movie review classification. It has two classes: positive and negative. Experiments are conducted on the first 500 sentences of the validation set.
\item \textbf{AG News}: A collection of news articles belonging to 4 different classes including World, Sports, Science/Technology and Business. Experiments are conducted on first 500 sentences of test set.
\item \textbf{IMDB}: IMDB website dataset for binary sentiment classification containing a set of highly polar movie reviews. Experiments are conducted on first 500 sentences of test set except for \victimL where only 100 sentences are used due to very high computation time due to very long average sentence length.
\end{itemize} 

\subsection{Interpretability Parameters}
\label{sec:interpretparams}
\paragraph{IG:} As integrated gradients is a gradient based approach and requires a reference baseline, we compute the attributions on the embedding space and set the reference baseline to the special token $<$PAD$>$ which is reserved in transformers as a special character. The step size for Integrated Gradients were chosen as 50 i.e. from reference to baseline, the gradients were summed up in 50 continuous steps.
\paragraph{LIME:} The number of perturbations for LIME were chosen as 500 and the maximum number of top-k words were chosen as 512 words - the truncation limit for all the models.

\subsection{Perturbation Parameters}
\label{sec:perturbparams}
We choose the number of nearest neighbours as 50 for swapping the words to limit the number of candidates. The maximum embedding cosine similarity between sentences was set as 0.5 to ensure sentences do not lose their semantic meaning. 

\subsection{Under the Hood}
\label{sec:implementdetails}

\paragraph{Pre-processing:} All sentences with less than 2 words in all datasets are removed due to word perturbations not existing in some cases. In other cases, the smaller sentence have a very big difference in rank correlation which can spuriously decrease evaluation metrics. Each sentence from all datasets is also converted to lower-case.

\begin{figure}[thb]
    \centering
     \includegraphics[width=0.45\textwidth]{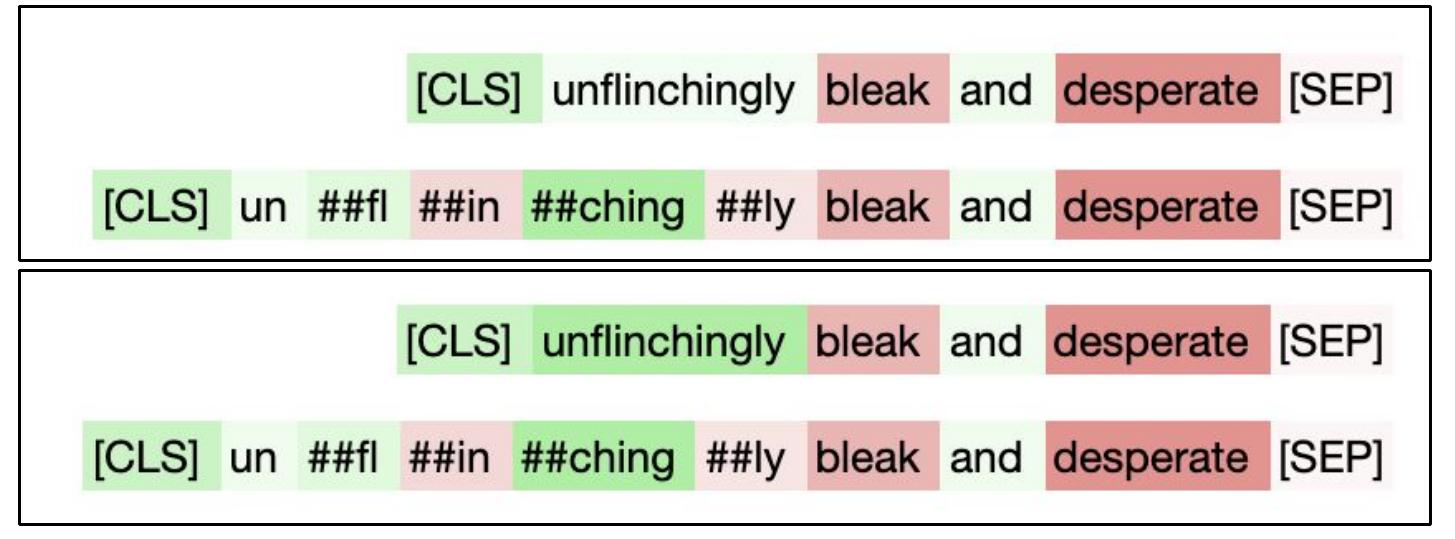} 
        \caption{The figure demonstrates the combining of tokens of a sentence tokenized using DistilBERT's pre-trained tokenizer. The top group of sentences demonstrates averaging approach and the bottom group of sentences are combined using Abs-Max approach detailed in \ref{sec:implementdetails}. [Best viewed in color]
          \label{fig:combine} }
\end{figure}

\paragraph{Fixing Tokenizations:} As pre-trained tokenizers for transformer models contain a ML matching based lookup vocabulary, many words in candidate sentences are tokenized in an unexpected manner. This results in the change of length of the token list which in turn changes the length of interpretations. To alleviate this problem, we test 2 distinct approaches to combine the unnaturally tokenized words into their original form.
\begin{itemize}
  \setlength\itemsep{0em}
    \item Average: The first approach combines all the tokens prefixed by a set character (\#\# in case of DistilBERT) into one single word and assigns the average value of the tokens to the combined tokens
    \item Max: The second approach combines all the tokens prefixed by a set character (\#\# in case of DistilBERT) into one single word and assigns the absolute maximum value with sign to the combined word.
\end{itemize}
Upon careful review, we utilize the second approach for our experiments. This is because, in uncommon cases where tokens hold opposite polarity to the ones in the word result in `diluted' value of the original token. An example of the effectiveness of the `Max' approach is given in Figure \ref{fig:combine}.

%% file: 05-evaluation.tex
\subsection{Evaluation Metrics} 
\label{sec:eval-metric}

\subsubsection{Rank Correlation}
To compare the correlation between interpretations of 2 sentences, we use the Spearman rank correlation metric. The more the ranks of the interpretations agree with each other, the higher the rank correlations. Importantly, we clip the negative values of the metric to 0. This is done because a negative correlation does not make sense when only comparing the difference in ranks and can spuriously bring down the average scores.  
\begin{equation}
    R-Correlation = \max (0, Spearman(I_1,I_2))
\end{equation}
We report results in Tables~\ref{tab:sst2-IG-rank},\ref{tab:sst2-LIME-rank},\ref{tab:agnews-IG-rank},\ref{tab:agnews-LIME-rank} and corresponding violin graphs  Figures~\ref{fig:violinplots-sst-ig-lom},\ref{fig:violinplots-sst-lime-l2} of average Spearman rank order correlations and standard deviations versus ratio of words perturbed for 3 datasets (SST-2, AG News and IMDB) across both  models (DistilBERT-uncased and RoBERTa-base) using 2 interpretability methods - \victimI and \victimL.

\subsubsection{Top-50\% Intersection}
To compare the extent to which the words with highest attributions are correctly predicted by both the interpretation methods, we use the Top-k\% intersection metric. To compute the intersection, we first find the words with the maximum absolute value of attributions (most important for prediction). We calculate the intersection of the top 50\% highest attribution words.
\begin{equation}
    Intersection = \frac{ \bigcap (argsort(I_1),argsort(I_2))}{0.5*length(I_1)}
\end{equation}
where argsort returns the indices of the top-50\% of the words in a sentence with highest attributions.

\subsubsection{Candidate Quality}
To judge the quality of the candidates generated using \method, we calculate two different commonly used quality metrics from adversarial attack literature - Perplexity and absolute number of grammar errors similar to \cite{li2020contextualized}.

\paragraph{Perplexity} We first use perplexity to estimate the fluency of candidates generated using \method. The lower the value, the more fluent the candidates, measured using a small size GPT-2 model (50k vocabulary) \cite{Radford2019LanguageMA}.

\paragraph{Grammatical Errors} Estimates the average number of absolute difference in grammatical errors between the original and the candidate sentences. We use the Language Tool \cite{naber2003rule} to compute the errors. 

%% file: 060-table-main.tex
\begin{table*}
\centering
\resizebox{0.95\textwidth}{!}{%
\begin{tabular}{c|ccc|ccc|ccc}
\hline
\multicolumn{10}{c}{SST-2} \\
\hline
\multirow{2}{*}{ } & \multicolumn{3}{c}{DistilBERT} & \multicolumn{3}{c}{RoBERTa} & \multicolumn{3}{c}{BERT-adv} \\
\hline
Ratio & L2 & $\Delta$LOM & Random & L2 & $\Delta$LOM & Random & L2 & $\Delta$LOM & Random \\
\hline
0-0.1 & 0.65 & 0.78 & 0.8 & 0.64 & 0.76 & 0.81 & 0.53 & 0.6 & 0.73 \\
0.1-0.2 & 0.53 & 0.65 & 0.64 & 0.57 & 0.61 & 0.69 & 0.43 & 0.43 & 0.52 \\
0.2-0.3 & 0.42 & 0.55 & 0.59 & 0.51 & 0.59 & 0.6 & 0.3 & 0.33 & 0.42 \\
0.3-0.4 & 0.36 & 0.48 & 0.48 & 0.47 & 0.47 & 0.55 & 0.35 & 0.3 & 0.43 \\
0.4-0.5 & 0.31 & 0.42 & 0.47 & 0.42 & 0.43 & 0.48 & 0.14 & 0.24 & 0.36 \\
\hline
\end{tabular}
}
\caption{Change in average rank-order correlation using metrics - \diffs, \nobjs and random selection conmputed using the interpretability method: \victimI, for dataset- SST-2 over 3 models - DistilBERT, RoBERTa and BERT-adv. \label{tab:sst2-IG-rank}
}
\resizebox{0.95\textwidth}{!}{%
\begin{tabular}{c|ccc|ccc|ccc}
\hline
\multicolumn{10}{c}{SST-2} \\
\hline
\multirow{2}{*}{ } & \multicolumn{3}{c}{DistilBERT} &
\multicolumn{3}{c}{RoBERTa} & \multicolumn{3}{c}{BERT-adv} \\
\hline
Ratio & L2 & $\Delta$LOM & Random & L2 & $\Delta$LOM & Random & L2 & $\Delta$LOM & Random \\
\hline
0-0.1 & 0.77 & 0.78 & 0.81 & 0.75 & 0.76 & 0.81 & 0.75 & 0.76 & 0.79 \\
0.1-0.2 & 0.71 & 0.71 & 0.73 & 0.71 & 0.71 & 0.74 & 0.68 & 0.68 & 0.7 \\
0.2-0.3 & 0.67 & 0.68 & 0.68 & 0.68 & 0.69 & 0.7 & 0.63 & 0.64 & 0.65 \\
0.3-0.4 & 0.65 & 0.65 & 0.65 & 0.66 & 0.67 & 0.67 & 0.61 & 0.61 & 0.64 \\
0.4-0.5 & 0.6 & 0.62 & 0.62 & 0.63 & 0.63 & 0.65 & 0.59 & 0.56 & 0.63 \\
\hline
\end{tabular}
}
\caption{Change in average Top-50\% intersection using metrics - \diffs, \nobjs and random selection conmputed using the interpretability method: \victimI, for dataset- SST-2 over 3 models - DistilBERT, RoBERTa and BERT-adv. \label{tab:sst2-IG-topk}}

\resizebox{0.95\textwidth}{!}{%
\begin{tabular}{c|ccc|ccc|ccc}
\hline
\multicolumn{10}{c}{SST-2} \\
\hline
& \multicolumn{3}{c}{DistilBERT} & \multicolumn{3}{c}{RoBERTa} & \multicolumn{3}{c}{BERT-adv} \\
\hline
Ratio & L2 & $\Delta$LOM & Random & L2 & $\Delta$LOM & Random & L2 & $\Delta$LOM & Random \\
\hline
0-0.1 & 0.64 & 0.7 & 0.79 & 0.59 & 0.66 & 0.76 & 0.57 & 0.68 & 0.72 \\
0.1-0.2 & 0.52 & 0.58 & 0.65 & 0.58 & 0.63 & 0.7 & 0.37 & 0.52 & 0.59 \\
0.2-0.3 & 0.46 & 0.51 & 0.56 & 0.52 & 0.58 & 0.62 & 0.34 & 0.47 & 0.54 \\
0.3-0.4 & 0.39 & 0.43 & 0.46 & 0.48 & 0.54 & 0.58 & 0.31 & 0.36 & 0.36 \\
0.4-0.5 & 0.23 & 0.29 & 0.46 & 0.55 & 0.55 & 0.54 & 0.28 & 0.2 & 0.24 \\
\hline
\end{tabular}
}
\caption{Change in average rank-order correlation using metrics - \diffs, \nobjs and random selection conmputed using the interpretability method: \victimL, for dataset- SST-2 over 3 models - DistilBERT, RoBERTa and BERT-adv. \label{tab:sst2-LIME-rank}}

\resizebox{0.95\textwidth}{!}{%
\begin{tabular}{c|ccc|ccc|ccc}
\hline
\multicolumn{10}{c}{SST-2} \\
\hline
& \multicolumn{3}{c}{DistilBERT} & \multicolumn{3}{c}{RoBERTa} & \multicolumn{3}{c}{BERT-adv} \\
\hline
Ratio & L2 & $\Delta$LOM & Random & L2 & $\Delta$LOM & Random & L2 & $\Delta$LOM & Random \\
\hline
0-0.1 & 0.64 & 0.7 & 0.79 & 0.59 & 0.66 & 0.76 & 0.57 & 0.68 & 0.72 \\
0.1-0.2 & 0.52 & 0.58 & 0.65 & 0.58 & 0.63 & 0.7 & 0.37 & 0.52 & 0.59 \\
0.2-0.3 & 0.46 & 0.51 & 0.56 & 0.52 & 0.58 & 0.62 & 0.34 & 0.47 & 0.54 \\
0.3-0.4 & 0.39 & 0.43 & 0.46 & 0.48 & 0.54 & 0.58 & 0.31 & 0.36 & 0.36 \\
0.4-0.5 & 0.23 & 0.29 & 0.46 & 0.55 & 0.55 & 0.54 & 0.28 & 0.2 & 0.24 \\
\hline
\end{tabular}
}
\caption{Change in average Top-50\% intersection using metrics - \diffs, \nobjs and random selection conmputed using the interpretability method: \victimL, for dataset- SST-2 over 3 models - DistilBERT, RoBERTa and BERT-adv. \label{tab:sst2-LIME-topk}}

\end{table*}

\begin{table*}
\centering
\resizebox{\textwidth}{!}{
\begin{tabular}{c|ccc|ccc|ccc}
\hline
\multicolumn{10}{c}{\textbf{Perplexity} (lower is better)} \\
\hline
& \multicolumn{3}{c}{DistilBERT} & \multicolumn{3}{c}{RoBERTa} & \multicolumn{3}{c}{BERT-adv} \\
\hline
Dataset & C-avg & $\Delta$LOM & L2 & C-avg & $\Delta$LOM & L2 & C-avg & $\Delta$LOM & L2 \\
\hline
SST-2 (130.85) & 352.55 & \textbf{285.88} & 286.62 & 272.67 & \textbf{245.55} & 248.86 & 388.99 & \textbf{237.76} & 238.14 \\
AGNews (76.18) & 359.13 & \textbf{239.95} & 241.33 & 275.31 & \textbf{194.89} & 195.29 & 352.05 & 230.44 & \textbf{229.38} \\
IMDB (39.12) & 101.71 & \textbf{65.04} & 65.1 & 101.3 & \textbf{62.41} & 63.65 & 84.51 & \textbf{63.21} & 64.58 \\
\hline
\end{tabular}
}
\caption{Average values of perplexity calculated using a small GPT-2 model over all candidates generated by \method (C-avg). The values in columns LOM and L2 denote the perplexity values calculated on the selected sentences using the proposed metrics. The average value of perplexity of original sentences in dataset are given in parentheses. Selection using metrics give more fluent sentences.}
\label{tab:ppl}
\end{table*}

%% file: 04-experiments.tex
\subsection{Model Choices}
The robustness concern of interpretation strategies challenges their use in critical applications, raising concerns like lack of trust. However, it is unclear what causes the ``fragile explanations'', the model or the interpretation?  We therefore select three different transformer models namely, DistilBERT-uncased \cite{sanh2019distilbert}, RoBERTa-base \cite{Liu2019RoBERTaAR} and BERT-base \cite{devlin2018bert} to conduct our experiments. 
More importantly, we retrain the BERT-base to obtain the BERT-base-adv model that is an adversarially trained version of the BERT-base model. The rationale behind the choices is to investigate the impact of model's robustness on the robustness of the interpretations. (1)~First, a generic transformer model like DistilBERT is relatively smaller and faster but less robust than the other two. (2)~Next RoBERTa is extensively better pre-trained and has a far more robust performance. (3) Lastly, BERT-base-adv model is trained from adversarial training. We use the popular TextFooler\cite{Jin_Jin_Zhou_Szolovits_2020} algorithm to generate adversarial examples via the open-source package Textattack.
DistilBERT and RoBERTa models were from pre-trained models, fine-tuned on the respective datasets and we take them from the Huggingface's transformer model hub\cite{wolf-etal-2020-transformers} without change. Differently, BERT-base-adv model is adversarially trained by attacking 10000 training examples for the IMDB and AG datasets and attacking all training samples for SST-2 dataset.



%% file: 06-results.tex
\section{Empirical Results}
\label{sec:results}

\begin{table}[t]
\centering
\resizebox{0.45\textwidth}{!}{
\begin{tabular}{c|c|ccccc}
\hline
 &  & \multicolumn{5}{c}{Number of Words Perturbed} \\
\hline
Dataset & Model & 0 & 1 & 2 & 3 & 4 \\
\hline
\multirow{3}{*}{ SST-2 } & DistilBERT & 0.97 & 0.95 & 0.92 & 0.88 & 0.82 \\
& RoBERTa & 0.98 & 0.98 & 0.98 & 0.98 & 0.98 \\
& BERT-adv & 0.97 & 0.96 & 0.94 & 0.92 & 0.91 \\
\hline
\multirow{3}{*}{ AGNews } & DistilBERT & 0.98 & 0.97 & 0.93 & 0.86 & 0.82 \\
& RoBERTa & 0.98 & 0.98 & 0.98 & 0.98 & 0.98 \\
& BERT-adv & 0.97 & 0.95 & 0.95 & 0.94 & 0.91 \\
\hline
\multirow{3}{*}{ IMDB } & DistilBERT & 0.99 & 0.97 & 0.94 & 0.89 & 0.84 \\
& RoBERTa & 0.98 & 0.98 & 0.98 & 0.98 & 0.98 \\
& BERT-adv & 0.97 & 0.96 & 0.94 & 0.94 & 0.91 \\
\hline
\end{tabular}
}
\caption{Average model confidence for correct prediction values for increasing number of words perturbed over models - DistilBERT, RoBERTa and BERT-adv on datasets - SST-2, AGNews and IMDB }
\label{tab:model-confidence}
\vspace{1mm}
\end{table}

\subsection{Rank Order and Top-50\% Intersection}
The results are reported in a tabular manner across 3 datasets (SST-2, AG News and IMDB), 3 models (DistilBERT, RoBERTa and BERT-adv~(Section~\ref{sec:appendix}(Appendix)) and 2 interpretability methods covering both metrics - \diffs, \dnobjs and compared against random candidate selection independent of both metrics. The first set of tables (Tables~\ref{tab:sst2-IG-rank} and \ref{tab:sst2-LIME-rank}) report the average rank-order correlation between interpretations from the perturbed and the original, across different perturbation ratios in buckets of 10\%. The second set of tables  (Tables~\ref{tab:sst2-IG-topk} and \ref{tab:sst2-LIME-topk}) report the average top-50\% intersection. The rank correlation results for the IMDB datasets are reported only on IG due to excessive computational constraints. Due to space constraints, the results for both AGNews and IMDB datasets are reported in (Tables~\ref{tab:agnews-IG-rank}-\ref{tab:agnews-LIME-topk} and Tables~\ref{tab:imdb-IG-rank}-\ref{tab:imdb-IG-topk} respectively, Section~\ref{sec:table2appx}~(Appendix)) along with a more detailed representation of the intra-bucket distribution in the form of Violin Graphs (Section~\ref{sec:violin2appx}~(Appendix)).\\ 
A bucket represents all instances of perturbed candidates in the ratio between that lower and higher range. For example, bucket between ``0.1-0.2'' contains all rank-order correlations from sentences with a percentage of words perturbed between 10\% and 20\%. We also provide violin plots in appendix showcasing intra-bucket distribution for the dataset SST-2 (Figures~\ref{fig:violinplots-sst-ig-lom}-\ref{fig:violinplots-sst-lime-l2}). We observe that both average rank-order correlation and top-50\% intersection scores decrease as the ratio of words being perturbed increases. Observations imply that interpretations of sentences become increasingly dissimilar to the original sentence as more words are perturbed even though the prediction robustness of the models remains high (see Table~\ref{tab:model-confidence}, Figure~\ref{fig:eg-sst2-distil-roberta}). Similar trends are observed across all models, datasets, and covering both victim interpretability methods. These empirical observations demonstrate interpretations generated by \victimI and \victimL are fragile for all models - even models that are adversarially more robust (BERT-adv). To further demonstrate effectiveness of proposed metrics, we plot violin plots on SST-2 dataset for avg. rank correlation versus selection using metrics and random. (Figure~\ref{fig:ablation-sst2} - Appendix)

%% file: 064-quality.tex
\subsection{Quality of candidates}
\paragraph{Perplexity} The average perplexity values over all models and datasets are reported in Table~\ref{tab:ppl}. For each dataset, model pair values corresponding to proposed metrics and random selection are reported.  It can be observed that perplexity of candidates selected using proposed metrics have lower perplexity score (implying better fluency) than average of all candidates generated by \method.

\paragraph{Grammatical Errors} Estimates average number of absolute difference in grammatical errors between the original and candidate sentences. We use Language Tool \cite{naber2003rule} to compute the errors. The results for SST-2 dataset are reported in Table~\ref{tab:grm}.


%% file: 065-quality-tables.tex
\begin{table}[t]
\centering
\resizebox{0.45\textwidth}{!}{
\begin{tabular}{c|ccc}
\hline
\multicolumn{4}{c}{Grammatical Errors (lower is better)} \\
\hline
Model & C-avg & L2 & $\Delta$LOM \\
\hline
DistilBERT & 0.59 & 0.59 & \textbf{0.58} \\
RoBERTa & 0.79 & 0.76 & \textbf{0.75} \\
BERT-adv & 0.60 & \textbf{0.51} & 0.52 \\
\hline
\end{tabular}
}
\caption{Average number of grammatical errors on candidates generated using \method on the SST-2 dataset (C-avg). The accompanying values in columns $\Delta$LOM and L2 denote the grammar errors calculated on the sentences selected using the proposed metrics.}
\label{tab:grm}
\vspace{1mm}
\end{table}


%% file: 07-conclusions.tex
\section{Conclusions}
\vspace{-2mm}
Literature sees a growing emphasis on interpretation techniques for explaining NLP model predictions. Our work demonstrates a novel algorithm that generates perturbed inputs that provide evidence of fragile interpretations. We demonstrate the effectiveness of our approach across three different models, with one of them adversarially trained. Our results show that it is possible to attack interpretations using simple input-level word swaps under certain constraints. We also demonstrate that both black and white-box interpretability approaches (\victimL and \victimI) show fragility in their derived interpretations. We hope our findings can pave lights for future studies on defending against problem of fragile interpretations in NLP.

%% file: 062-baseline2appx.tex
\subsection{Compare with Baseline}
\label{sec:ablation2appx}
\begin{figure}[t]
    \centering
     \includegraphics[width=0.4\textwidth]{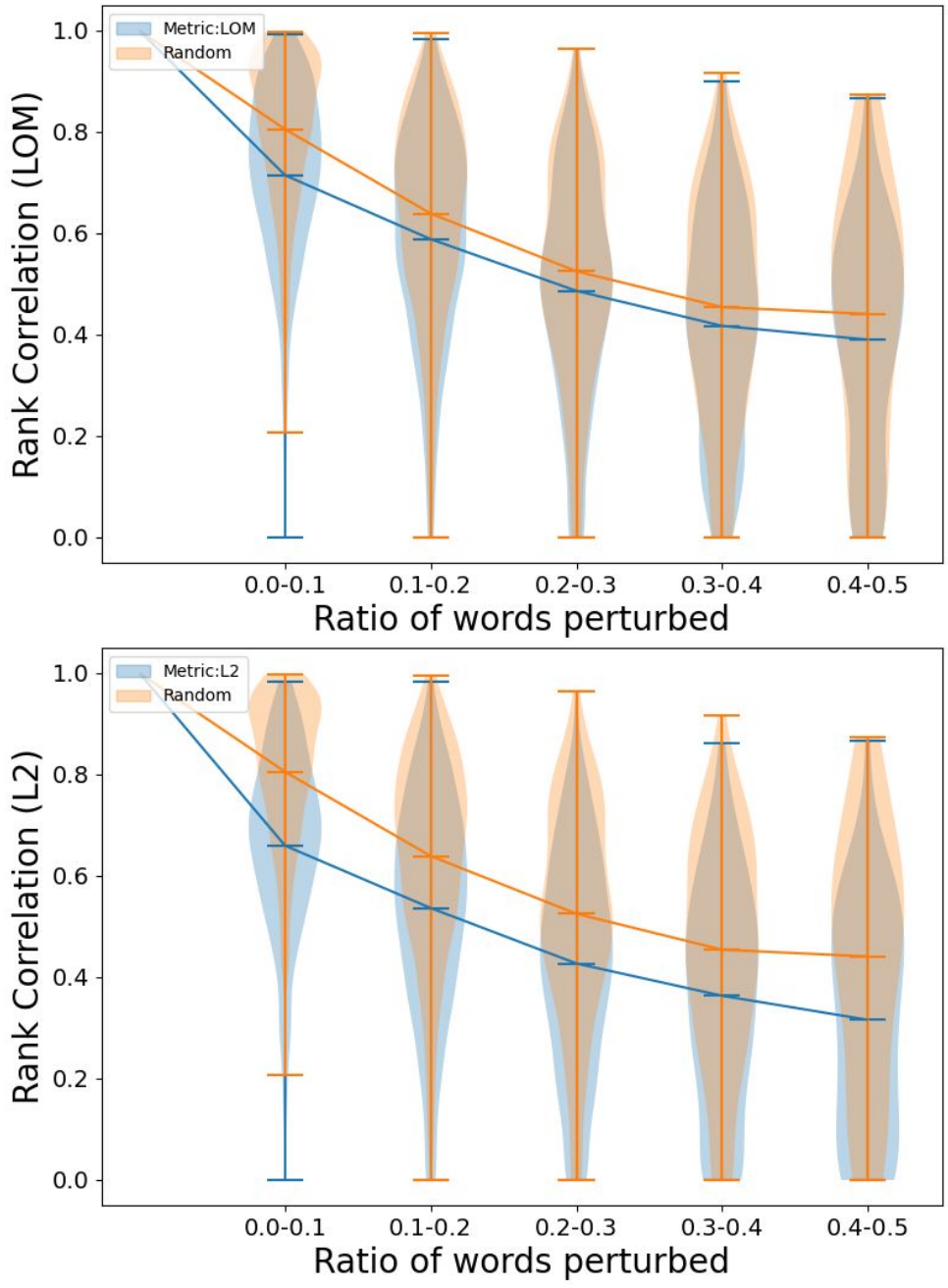} 
        \caption{The violin graphs demonstrate the effectiveness of candidate selection based on the proposed metrics \nobjs and \diffs over random selection for SST-2 dataset. As it can be seen that the selection based on the proposed metrics disrupts rank correlation more as compared to randomly selecting candidates.
          \label{fig:ablation-sst2} }
\end{figure}

\begin{figure}[t]
    \centering
     \includegraphics[width=0.4\textwidth]{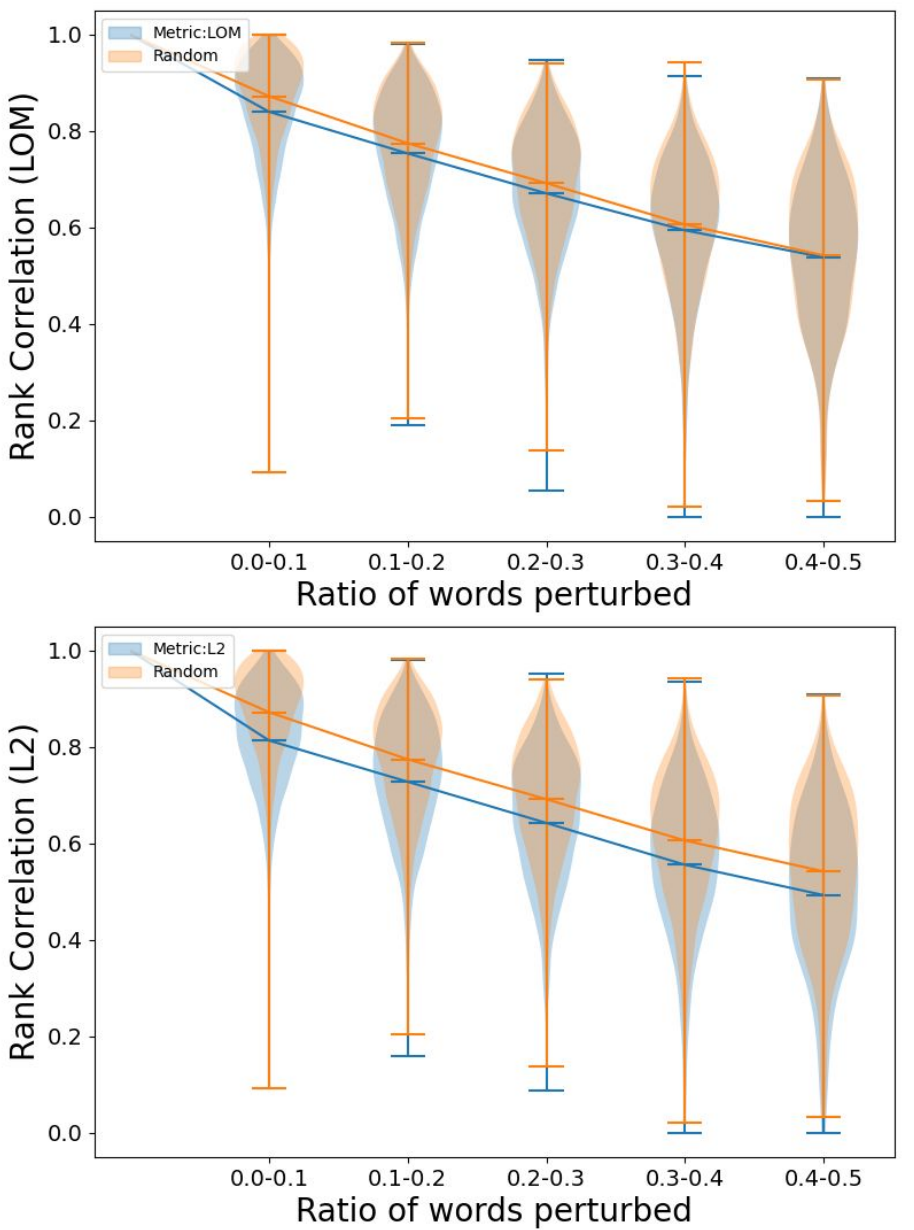} 
        \caption{The violin graphs demonstrate the effectiveness of candidate selection based on the proposed metrics \nobjs and \diffs over random selection for AGNews dataset. As it can be seen that the selection based on the proposed metrics disrupts rank correlation more as compared to randomly selecting candidates.
          \label{fig:ablation-agnews} }
\end{figure}

Figures~\ref{fig:ablation-sst2},\ref{fig:ablation-agnews} show the decrease in average rank correlation when considering random candidates as opposed to selection using the \nobjs metric.

\subsection{Additional Results}
\label{sec:table2appx}
In this section we report the average rank order correlation and the average top-50\% intersection scores for AGNews and IMDB datasets. The Tables~\ref{tab:agnews-IG-rank},\ref{tab:agnews-IG-topk} correspond to AGNews' rank correlation and top-50\% scores using \victimI whereas Tables~\ref{tab:agnews-LIME-rank},\ref{tab:agnews-LIME-topk} show same values using \victimL. Tables~\ref{tab:imdb-IG-rank} and \ref{tab:imdb-IG-topk} show similar values but for IMDB dataset.


\subsection{Violin Plots for intra-bucket distribution analysis}
\label{sec:violin2appx}
The Violin plots convey more information about the relative distribution of average rank correlations and Top-50\% values for various bucket ratios. The following figures are only reported on the SST-2 dataset for each combination of evaluation metric and interpretability methods.

\subsection{Visual Results}
\label{sec:visual-res}
A few visual results demonstrating the gradual change in interpretations of candidate adversaries is shown in Figure \ref{fig:eg-sst2-distil-roberta}. It can be observed that $\Delta$LOM score gradually increases with word perturbations. The examples demonstrate the same 3 sentences from the dataset perturbed under DistilBERT and RoBERTa respectively.

%% file: 063-tables2Appx.tex

\begin{table*}
\centering
\resizebox{0.85\textwidth}{!}{%
\begin{tabular}{c|ccc|ccc|ccc}
\hline
\multicolumn{10}{c}{AGNews} \\
\hline
& \multicolumn{3}{c}{DistilBERT} & \multicolumn{3}{c}{RoBERTa} & \multicolumn{3}{c}{BERT-adv} \\
\hline
Ratio & L2 & $\Delta$LOM & Random & L2 & $\Delta$LOM & Random & L2 & $\Delta$LOM & Random \\
\hline
0-0.1 & 0.81 & 0.84 & 0.86 & 0.73 & 0.68 & 0.82 & 0.38 & 0.56 & 0.63 \\
0.1-0.2 & 0.72 & 0.75 & 0.78 & 0.65 & 0.57 & 0.72 & 0.32 & 0.42 & 0.46 \\
0.2-0.3 & 0.64 & 0.66 & 0.69 & 0.62 & 0.52 & 0.66 & 0.28 & 0.32 & 0.29 \\
0.3-0.4 & 0.55 & 0.58 & 0.58 & 0.58 & 0.48 & 0.62 & 0.25 & 0.25 & 0.26 \\
0.4-0.5 & 0.49 & 0.52 & 0.56 & 0.52 & 0.42 & 0.56 & 0.18 & 0.23 & 0.24 \\
\hline
\end{tabular}
}
\caption{Change in average rank-order correlation using metrics - \diffs, \nobjs and random selection conmputed using the interpretability method: \victimI, for dataset- AGNews over 3 models - DistilBERT, RoBERTa and BERT-adv. \label{tab:agnews-IG-rank}
}
\resizebox{0.85\textwidth}{!}{%
\begin{tabular}{c|ccc|ccc|ccc}
\hline
\multicolumn{10}{c}{AGNews} \\
\hline
& \multicolumn{3}{c}{DistilBERT} & \multicolumn{3}{c}{RoBERTa} & \multicolumn{3}{c}{BERT-adv} \\
\hline
Ratio & L2 & $\Delta$LOM & Random & L2 & $\Delta$LOM & Random & L2 & $\Delta$LOM & Random \\
\hline
0-0.1 & 0.64 & 0.65 & 0.71 & 0.7 & 0.74 & 0.85 & 0.48 & 0.51 & 0.79 \\
0.1-0.2 & 0.57 & 0.58 & 0.69 & 0.61 & 0.64 & 0.8 & 0.37 & 0.4 & 0.69 \\
0.2-0.3 & 0.57 & 0.58 & 0.62 & 0.55 & 0.59 & 0.77 & 0.24 & 0.27 & 0.64 \\
0.3-0.4 & 0.53 & 0.53 & 0.58 & 0.52 & 0.55 & 0.74 & 0.22 & 0.24 & 0.6 \\
0.4-0.5 & 0.51 & 0.52 & 0.56 & 0.45 & 0.5 & 0.71 & 0.19 & 0.24 & 0.58 \\
\hline
\end{tabular}

}
\caption{Change in average Top-50\% intersection using metrics - \diffs, \nobjs and random selection conmputed using the interpretability method: \victimI, for dataset- AGNews over 3 models - DistilBERT, RoBERTa and BERT-adv. \label{tab:agnews-IG-topk}}

\resizebox{0.85\textwidth}{!}{%
\begin{tabular}{c|ccc|ccc|ccc}
\hline
\multicolumn{10}{c}{AGNews} \\
\hline
& \multicolumn{3}{c}{DistilBERT} & \multicolumn{3}{c}{RoBERTa} & \multicolumn{3}{c}{BERT-adv} \\
\hline
Ratio & L2 & $\Delta$LOM & Random & L2 & $\Delta$LOM & Random & L2 & $\Delta$LOM & Random \\
\hline
0-0.1 & 0.65 & 0.69 & 0.71 & 0.58 & 0.57 & 0.61 & 0.7 & 0.61 & 0.72 \\
0.1-0.2 & 0.59 & 0.6 & 0.62 & 0.55 & 0.54 & 0.56 & 0.69 & 0.45 & 0.7 \\
0.2-0.3 & 0.53 & 0.53 & 0.58 & 0.54 & 0.53 & 0.48 & 0.65 & 0.35 & 0.66 \\
0.3-0.4 & 0.48 & 0.52 & 0.55 & 0.51 & 0.51 & 0.36 & 0.65 & 0.28 & 0.65 \\
0.4-0.5 & 0.44 & 0.38 & 0.46 & 0.43 & 0.42 & 0.43 & 0.59 & 0.26 & 0.61 \\
\hline
\end{tabular}
}
\caption{Change in average rank-order correlation using metrics - \diffs, \nobjs and random selection conmputed using the interpretability method: \victimL, for dataset- AGNews over 3 models - DistilBERT, RoBERTa and BERT-adv. \label{tab:agnews-LIME-rank}}

\resizebox{0.85\textwidth}{!}{%
\begin{tabular}{c|ccc|ccc|ccc}
\hline
\multicolumn{10}{c}{AGNews} \\
\hline
& \multicolumn{3}{c}{DistilBERT} & \multicolumn{3}{c}{RoBERTa} & \multicolumn{3}{c}{BERT-adv} \\
\hline
Ratio & L2 & $\Delta$LOM & Random & L2 & $\Delta$LOM & Random & L2 & $\Delta$LOM & Random \\
\hline
0-0.1 & 0.62 & 0.64 & 0.66 & 0.6 & 0.62 & 0.61 & 0.56 & 0.56 & 0.55 \\
0.1-0.2 & 0.58 & 0.59 & 0.63 & 0.58 & 0.58 & 0.58 & 0.53 & 0.54 & 0.53 \\
0.2-0.3 & 0.57 & 0.57 & 0.58 & 0.55 & 0.57 & 0.57 & 0.51 & 0.51 & 0.52 \\
0.3-0.4 & 0.55 & 0.56 & 0.58 & 0.55 & 0.55 & 0.57 & 0.5 & 0.5 & 0.52 \\
0.4-0.5 & 0.53 & 0.55 & 0.57 & 0.54 & 0.54 & 0.56 & 0.51 & 0.5 & 0.52 \\
\hline
\end{tabular}
}
\caption{Change in average Top-50\% intersection using metrics - \diffs, \nobjs and random selection conmputed using the interpretability method: \victimL, for dataset- AGNews over 3 models - DistilBERT, RoBERTa and BERT-adv. \label{tab:agnews-LIME-topk}}
\end{table*}

\begin{table*}
\centering
\resizebox{0.95\textwidth}{!}{%
\begin{tabular}{c|ccc|ccc|ccc}
\hline
\multicolumn{10}{c}{IMDB} \\
\hline
& \multicolumn{3}{c}{DistilBERT} & \multicolumn{3}{c}{RoBERTa} & \multicolumn{3}{c}{BERT-adv} \\
\hline
Ratio & L2 & $\Delta$LOM & Random & L2 & $\Delta$LOM & Random & L2 & $\Delta$LOM & Random \\
\hline
0-0.1 & 0.69 & 0.69 & 0.71 & 0.75 & 0.74 & 0.8 & 0.45 & 0.44 & 0.55 \\
0.1-0.2 & 0.53 & 0.55 & 0.61 & 0.64 & 0.59 & 0.69 & 0.32 & 0.32 & 0.41 \\
0.2-0.3 & 0.41 & 0.44 & 0.5 & 0.51 & 0.48 & 0.58 & 0.28 & 0.29 & 0.39 \\
0.3-0.4 & 0.42 & 0.39 & 0.49 & 0.45 & 0.41 & 0.51 & 0.27 & 0.28 & 0.34 \\
0.4-0.5 & 0.33 & 0.31 & 0.41 & 0.34 & 0.37 & 0.49 & 0.12 & 0.17 & 0.28 \\
\hline
\end{tabular}
}
\caption{Change in average rank-order correlation using metrics - \diffs, \nobjs and random selection conmputed using the interpretability method: \victimI, for dataset- IMDB over 3 models - DistilBERT, RoBERTa and BERT-adv. \label{tab:imdb-IG-rank}}

\centering
\resizebox{0.95\textwidth}{!}{%
\begin{tabular}{c|ccc|ccc|ccc}
\hline
\multicolumn{10}{c}{IMDB} \\
\hline
& \multicolumn{3}{c}{DistilBERT} & \multicolumn{3}{c}{RoBERTa} & \multicolumn{3}{c}{BERT-adv} \\
\hline
Ratio & L2 & $\Delta$LOM & Random & L2 & $\Delta$LOM & Random & L2 & $\Delta$LOM & Random \\
\hline
0-0.1 & 0.7 & 0.71 & 0.74 & 0.73 & 0.75 & 0.76 & 0.61 & 0.63 & 0.66 \\
0.1-0.2 & 0.6 & 0.63 & 0.66 & 0.64 & 0.66 & 0.69 & 0.58 & 0.61 & 0.63 \\
0.2-0.3 & 0.59 & 0.6 & 0.63 & 0.57 & 0.63 & 0.65 & 0.57 & 0.6 & 0.61 \\
0.3-0.4 & 0.56 & 0.57 & 0.57 & 0.57 & 0.58 & 0.6 & 0.55 & 0.57 & 0.58 \\
0.4-0.5 & 0.52 & 0.52 & 0.52 & 0.52 & 0.55 & 0.57 & 0.54 & 0.54 & 0.54 \\
\hline
\end{tabular}
}
\caption{Change in average rank-order correlation using metrics - \diffs, \nobjs and random selection conmputed using the interpretability method: \victimI, for dataset- IMDB over 3 models - DistilBERT, RoBERTa and BERT-adv. \label{tab:imdb-IG-topk}}
\end{table*}

\clearpage

%% file: 063-violin2appx.tex
\begin{figure*}[t]
\centering 
    \minipage{0.33\textwidth}
     \includegraphics[width=\textwidth]{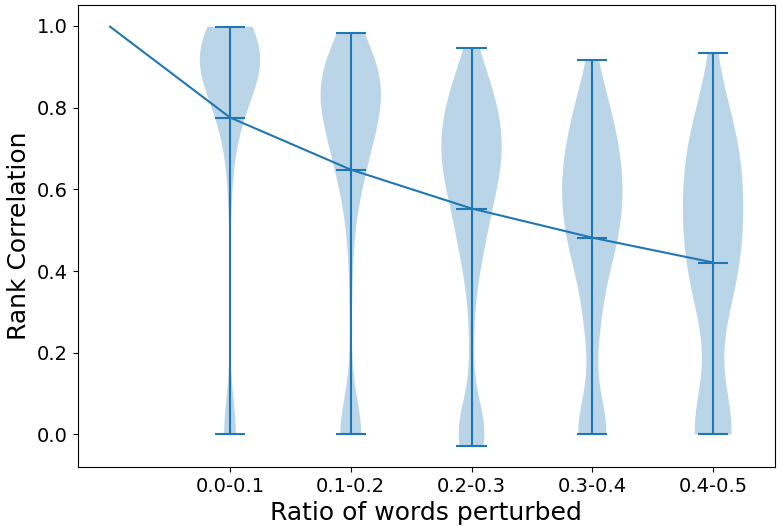}
    \endminipage\hfill
    \minipage{0.33\textwidth}
     \includegraphics[width=\textwidth]{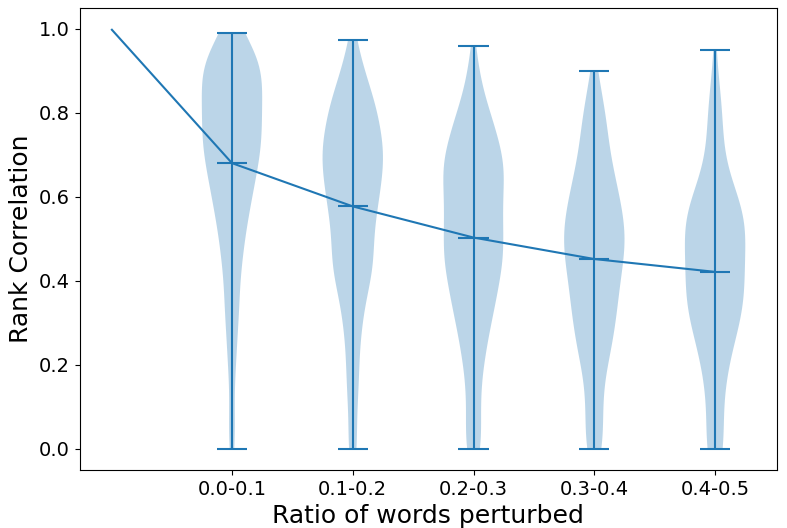} 
    \endminipage\hfill
    \minipage{0.34\textwidth}
     \includegraphics[width=\textwidth]{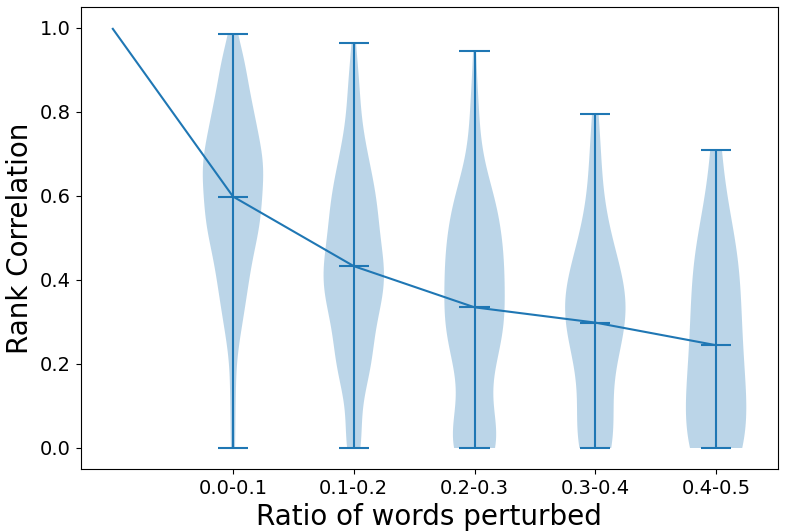} 
    \endminipage\hfill
    \caption{Average Rank-correlation for the dataset: SST-2, using metric: \nobjs on models DistilBERT, RoBERTa and BERT-adv using interpretability method -\victimI }
    \label{fig:violinplots-sst-ig-lom}
    \minipage{0.33\textwidth}
     \includegraphics[width=\textwidth]{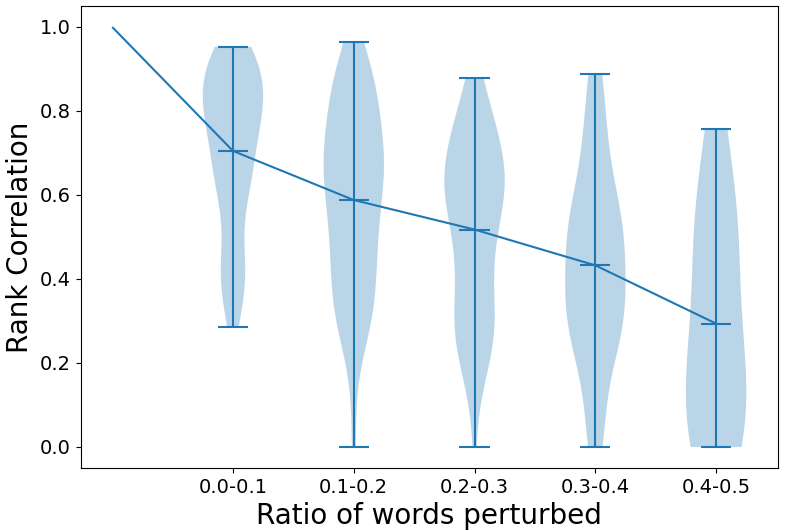} 
    \endminipage\hfill
    \minipage{0.33\textwidth}
     \includegraphics[width=\textwidth]{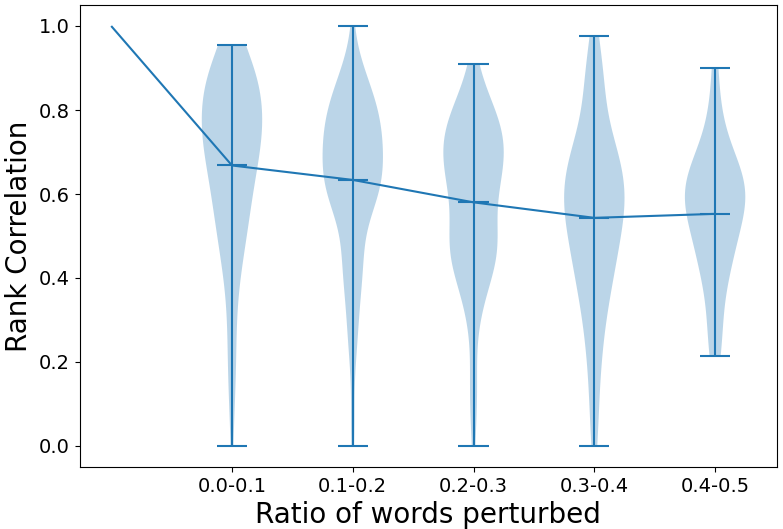} 
    \endminipage\hfill
    \minipage{0.33\textwidth}
     \includegraphics[width=\textwidth]{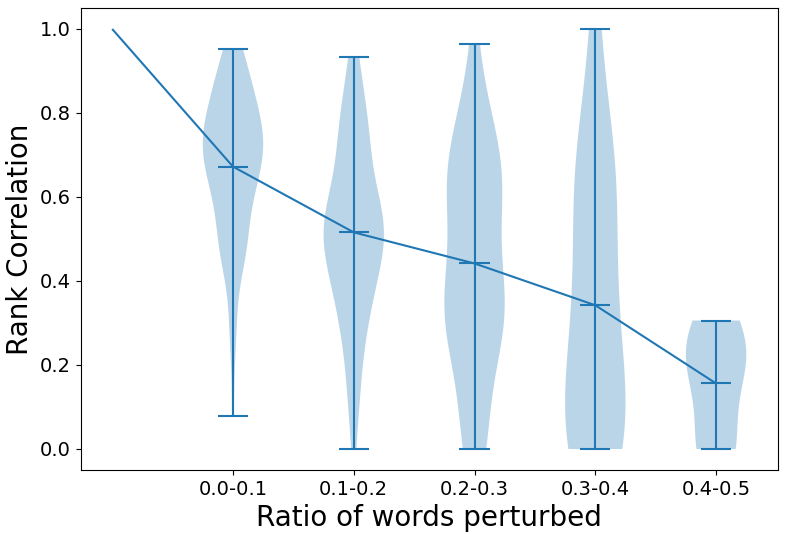} 
    \endminipage\hfill
    \caption{Average Rank-correlation for the dataset: SST-2, using metric: \nobjs on models DistilBERT, RoBERTa and BERT-adv using interpretability method -\victimL }
    \minipage{0.33\textwidth}
     \includegraphics[width=\textwidth]{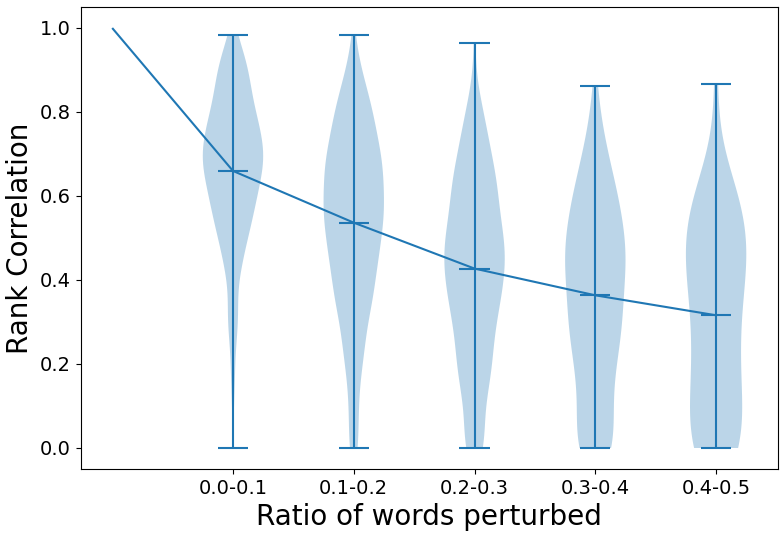}
    \endminipage\hfill
    \minipage{0.33\textwidth}
     \includegraphics[width=\textwidth]{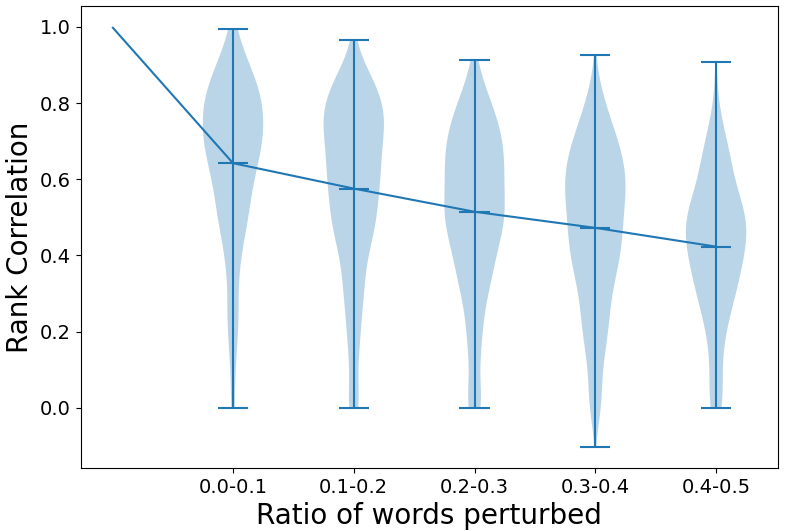} 
    \endminipage\hfill
    \minipage{0.33\textwidth}
     \includegraphics[width=\textwidth]{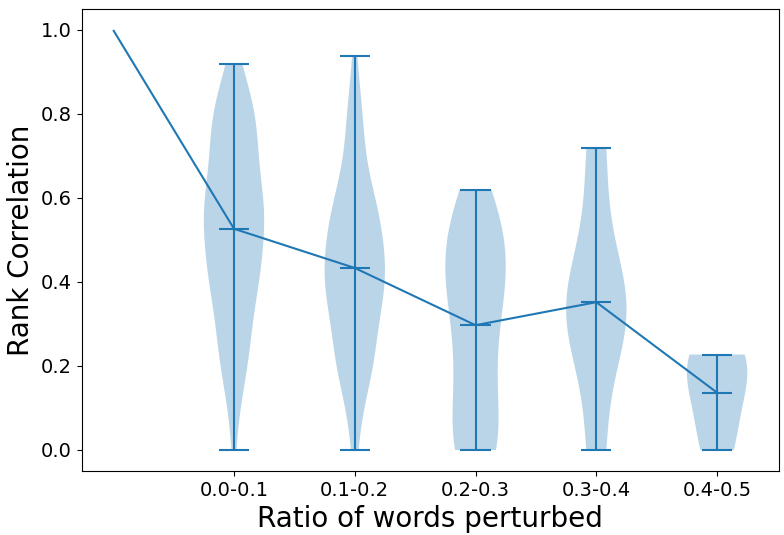} 
    \endminipage\hfill
    \caption{Average Rank-correlation for the dataset: SST-2, using metric: \diffs on models DistilBERT, RoBERTa and BERT-adv using interpretability method -\victimI }
    \minipage{0.33\textwidth}
     \includegraphics[width=\textwidth]{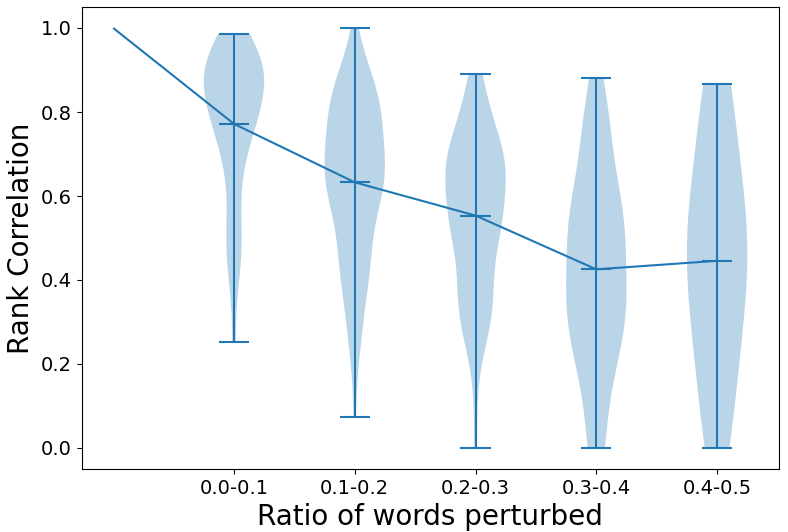} 
    \endminipage\hfill
    \minipage{0.33\textwidth}
     \includegraphics[width=\textwidth]{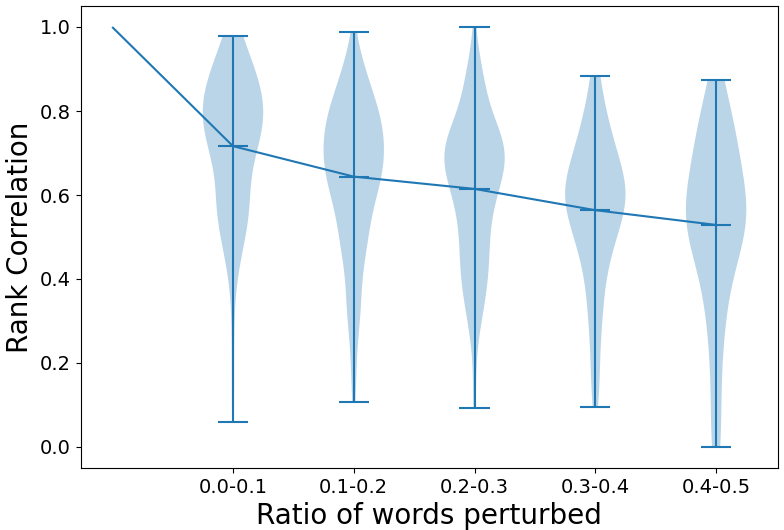} 
    \endminipage\hfill
    \minipage{0.33\textwidth}
     \includegraphics[width=\textwidth]{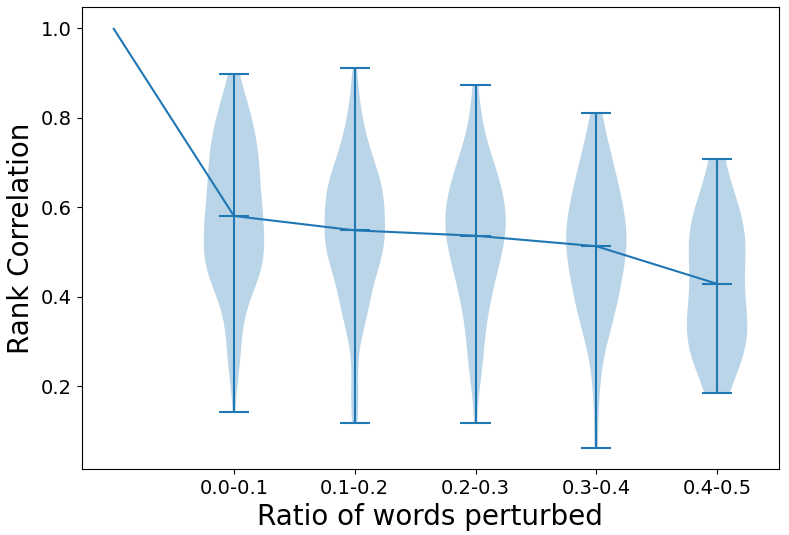}
    \endminipage\hfill
    \caption{Average Rank-correlation for the dataset: SST-2, using metric: \diffs on models DistilBERT, RoBERTa and BERT-adv using interpretability method -\victimL }
    \label{fig:violinplots-sst-lime-l2}
\end{figure*}

\clearpage

%% file: 99-visual.tex
\begin{figure*}[ht]
\centering 
    \begin{minipage}[h]{0.8\hsize} 
     \includegraphics[width=\textwidth]{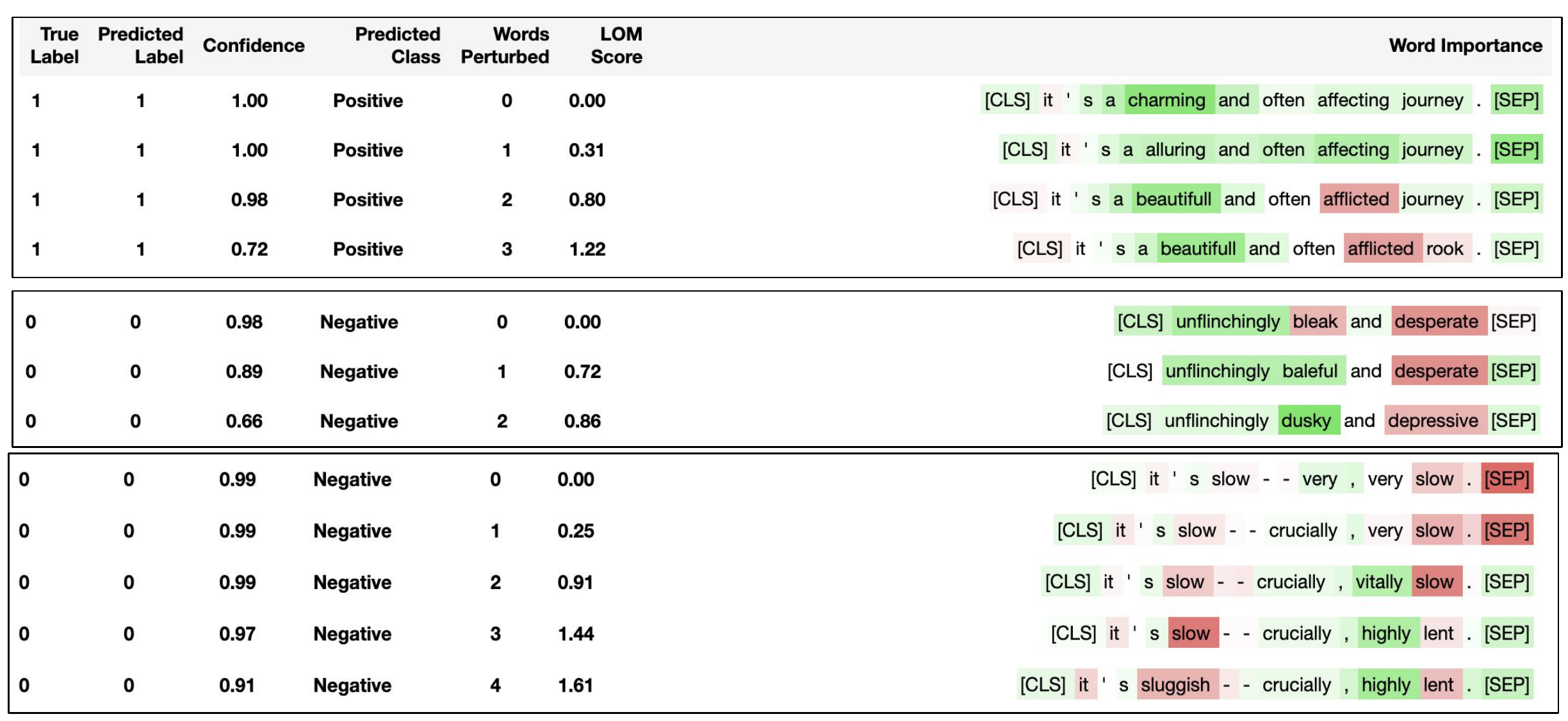} 
    \end{minipage} 
    \caption{A few random sentence explanations from the SST-2 dataset calculated on DistilBERT-uncased using \victimI. [Best viewed in color]. }
    \begin{minipage}[h]{0.8\hsize} 
     \includegraphics[width=\textwidth]{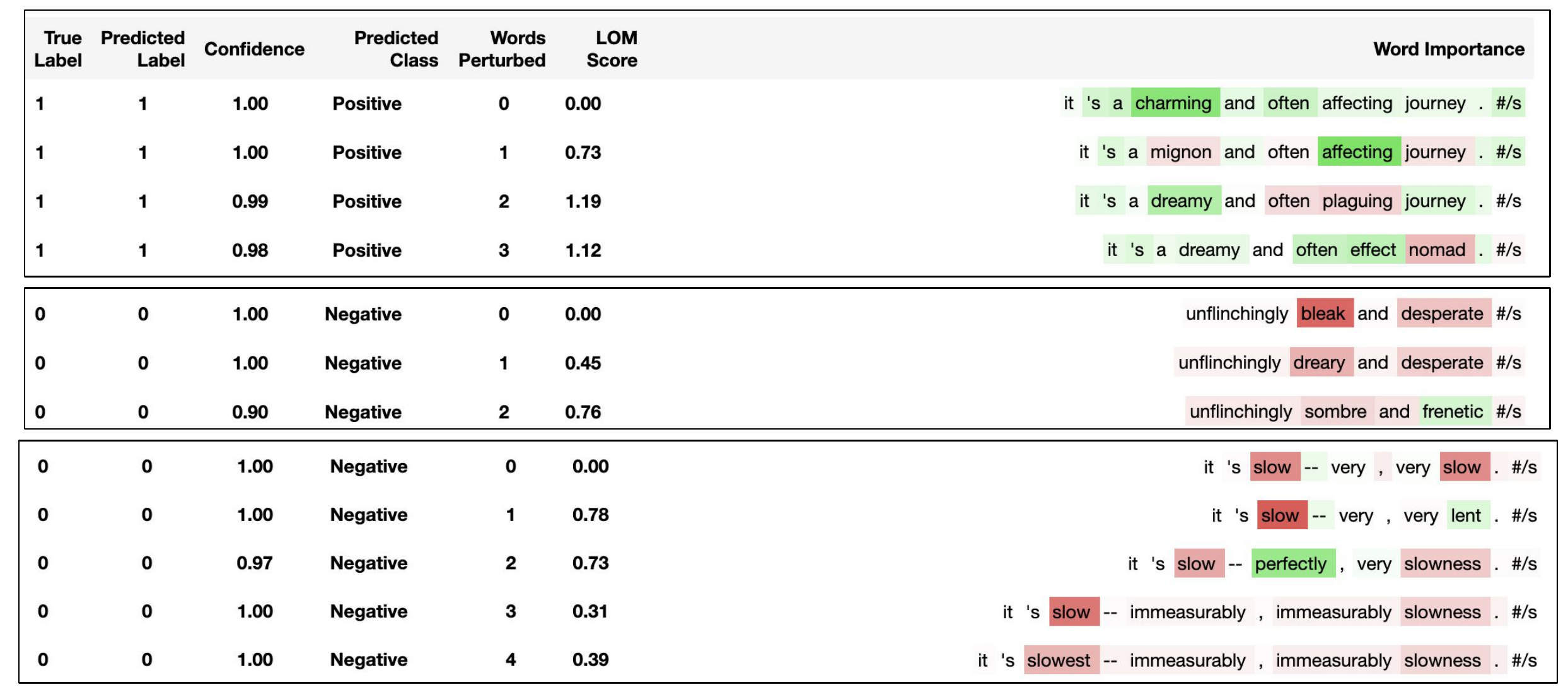} 
    \end{minipage} 
    \caption{The same sentence visualizations calculated on RoBERTa-base. It is clear RoBERTa is much more robust in making predictions but both DistilBERT and RoBERTa are susceptible to such attacks on their interpretations. [Best viewed in color] 
    \label{fig:eg-sst2-distil-roberta} }
\end{figure*}